\pdfoutput=1

\documentclass[11pt]{article}

\usepackage[]{acl}

\usepackage{times}
\usepackage{latexsym}

\usepackage[T1]{fontenc}

\usepackage[utf8]{inputenc}

\usepackage{microtype}

\usepackage{graphicx}
\usepackage{multirow}
\usepackage{hyperref}

\title{Predicting Desirable Revisions of Evidence and Reasoning in Argumentative Writing}

\author{Tazin Afrin \and Diane Litman\\
  University of Pittsburgh \\ Pittsburgh, Pennsylvania 15260 \\
  \texttt{\{taa74,dlitman\}@pitt.edu} \\}

\begin{document}
\maketitle
\begin{abstract}
 
We develop models to classify desirable evidence and desirable reasoning revisions in student argumentative writing. We explore two ways to improve classifier performance -- using the essay context of the revision, and using the feedback students received before the revision. 
We perform both intrinsic and extrinsic evaluation for each of our models and report a qualitative analysis. Our results show that while a model using  feedback information improves over a baseline model, models utilizing context - either alone or with feedback - are the most successful in identifying desirable revisions.

\end{abstract}

\section{Introduction}
\label{section: introduction}

\begin{table*}[!htb]
\centering
\begin{tabular}{|c|p{0.37\textwidth}|p{0.37\textwidth}|p{0.12\textwidth}|}
\hline			
\multicolumn{4}{|p{0.9\textwidth}|}{Feedback message: ``....Explain how the evidence helps to make your point... ... Tie the evidence not only to the point you are making within a paragraph, but to your overall argument..."}		\\
\hline \hline
& Original Draft	&	Revised Draft	&	Revision	\\	\hline \hline


1.& The author convinced me by saying in the passage that, "The plan is to get people out of poverty, assure them access to health care and help them stabilize the economy and quality of life in their communities."	&	The author convinced me by saying in the passage that, "The plan is to get people out of poverty, assure them access to health care and help them stabilize the economy and quality of life in their communities."	&	No-change	\\	\hline
2.	&  & ... & \\ \hline
	
3.&	&	They can do that by assuring that the people of Sauri, Kenya have food, water, liter, and a place to stay.	&	Added Desirable Reasoning	\\	\hline
4.& Also, in paragraph 3 it says, "The goals are supposed to be met by 2025; some other targets are set for 2035."	&	Also, in paragraph 3 it says, "The goals are supposed to be met by 2025; some other targets are set for 2035."	&	No-change	\\	\hline

5.& 	&	If the plans are going to be achieved in 2025 than their plans will be achieved in only 7 more years which would be in our life time.	&	Added Desirable Evidence	\\	\hline
6.	& ... & ... & \\ \hline

7.& Since so many people weren't fighting against poverty in 2010 people were being sent to the hospital and not even being treated cause they didn't have the money so, so many people died.	&	Since so many people weren't fighting against poverty in 2010 people were being sent to the hospital and not even being treated cause they didn't have the money so, so many people died.	&	No-change	\\	\hline

8.	& ... & ... & \\ \hline
9.& 	&	The kids and their families didn't have the money but but this supports my evidence by talking about how the kids don't go to school it's because them and their family are in poverty.	&	Added\quad Undesirable Reasoning	\\	\hline

\end{tabular}
\caption{Example of revisions extracted from an essay from our elementary-school dataset.}
\label{table: example revision}
\end{table*}

Successful essay writing by 
students typically involves multiple rounds of revision and assistance from 
teachers, peers, or automated writing evaluation (AWE) systems. Natural language processing (NLP) has become a key component of  AWE systems,  
with NLP being used to assess the content and structure of student writing and to automatically provide formative 
feedback~\cite{beigman2020AWE,zhang2016hl,ets-writing-mentor,wang2020eRevise}. 
While some students produce revised texts that are in line with the feedback automatically generated by a system or provided by other humans, 
other students either ignore the feedback or are unsuccessful in their feedback implementation attempts \cite{wang2020eRevise}.
Hence, analyzing student revisions in terms of their {\it desirability for improving essay quality} is important.
The development of AWE systems that leverage NLP to analyze a revision's alignment to feedback messages is one approach to convey to students a sense of a good revision direction.


Our research focuses on the {\it automatic classification of desirable and undesirable revisions of evidence use and reasoning}~\footnote{Such revisions of text {\it content} are generally considered most important in revising~\cite{faigley1981w}.} in argumentative writing. Argumentative writing is a skill that students need to develop to be strong writers and learners. 
By evidence use, we refer to examples and details that students use to support an argument. By reasoning, we refer to how  evidence is explained and linked  to an overall argument. Desirable revisions (e.g., add relevant evidence) are student revisions that have hypothesized utility in improving an essay in response to feedback (e.g., add more evidence), while undesirable revisions (e.g., add irrelevant evidence) do not have such hypothesized utility. 



Table~\ref{table: example revision} shows example desirable and undesirable revisions of evidence and reasoning from original to revised drafts of an essay aligned at the sentence-level. In response   to the feedback shown at the top of  Table~\ref{table: example revision}, 
 the student adds both reasoning and evidence.  Sentences 3, 5, and 9 are added desirable reasoning, desirable evidence, and undesirable reasoning respectively. 
The student also modified fluency in other sentences which is not shown here. Sentences 1, 4, and 7 are identical in both drafts.

In this paper, we first describe the labeling of desirable and undesirable revisions in three existing corpora of evidence and reasoning revisions. We then describe a baseline model and enhanced models using   context and feedback information to predict revision desirability. Finally, we present results from  intrinsic and extrinsic evaluations to demonstrate the utility of our enhanced models.
\section{Related Work}

NLP research on revision analysis primarily focuses on two 
domains: Wikipedia and academic writing. Studies in Wikipedia revisions focused on error correction, paraphrase or vandalism detection~\cite{daxenberger2012g}, factual versus fluency edits~\cite{bronner2012m}, semantic edit intention~\cite{yang2017hkh}, etc. 
In academic writing, revision studies have instead focused on defining revisions purpose tailored to argumentative writing ~\cite{zhang2015l,kashefi2022} 
and understanding the pattern of revisions~\cite{afrin2019EditorRole,shibani2018kb}. 
Exploring the pattern of iterative revision have also been studied in scientific writing~\cite{du-etal-2022-understanding-iterative}. 
While there have been some attempts at defining revisions in terms of their {\it quality} (e.g.,  vagueness of Wikipedia edits ~\cite{debnath-roth-2021-computational},  statement strength in scientific writing~\cite{tan2014l}, quality of claims in online debate~\cite{skitalinskaya-etal-2021-learning}, and improvement in argumentative writing ~\cite{afrin2018improvement}), they fail to incorporate feedback students were provided. 
~\citet{afrin2020RER} is the first study that touched on student revisions in terms of their utility in improving the essay with respect to automated feedback messages. However, their framework was applied to one dataset and they did not investigate state-of-the-art models for automatic classification. 
In this work, we focus on a simplified binary classification task to distinguish between desirable and undesirable revisions in student argumentative writing, and particularly explore the utility of two predictors of revision desirability - {\it context and feedback}. We also apply our model on {\it multiple student corpora}.


%
%
%
\begin{table*}[t]
\centering

\begin{tabular}{|l|c|c|c|c|c|c|}
\hline											
\multirow{2}{*}{Datasets}	&	\multirow{2}{*}{\#Students}	& Grade & Feedback	& 		Essay Drafts		&   Essay Score 	& Improvement \\
	&		&             Level                 & Source	&   Used	&  Range &  Score Range\\ \hline
Elementary	& 143   &	 $5^{th}$ \& $6^{th}$   & AWE	&   1 and 2	&   [1, 4]   &  [0, 3]\\ \hline
High-school	&	47  & 	  $12^{th}$             &  peer	&   1 and 2	&   [0, 5]  &  [-2, +3] \\ \hline
College	    &	60	&       college             & X	&   2 and 3	&   [15, 33]    &  -1, +1\\ \hline
\end{tabular}
\caption{Comparison of datasets used in this study (X = Not available).}
\label{table:datasets}
\end{table*}

\begin{table*}[!ht]
\centering
\begin{tabular}{|p{0.12\linewidth}|p{0.845\linewidth}|}

\hline	
\textbf{Data}	&	\textbf{Example Feedback}	\\ \hline 
Elementary (AWE 	&	Explain the evidence: 
 Tell your reader why you included each piece of evidence. Explain how the evidence helps to
make your point. \\
generated) & Explain how the evidence connects to the main idea \& elaborate:   Tie the evidence not only to the point you are making within a paragraph, but to your overall argument.  Elaborate. Give a detailed and clear explanation of how the evidence supports your argument.	\\ \hline
High-school (peer feedback)	&	for the spendthrifts and the hoaders, you used a good example for spendthifts but im confused on where you example for hoardering is. if it is mike tyson, i think you should include more detail about that. your fifth circle could use more detail as to what exactly made him hate man, because im confused about the story.	\\ \hline
\end{tabular}
\caption{Examples of feedback messages from elementary and high-school data.}
\label{table:feedback examples}
\end{table*}

Previous revision classification approaches 
either do not create contextual features ~\cite{daxenbergerG13g,zhang2015l}, or the context features represent only shallow information such as `location'~\cite{zhang2015l}. \citet{zhang2016l-context} incorporated context by using cohesion blocks focusing on adjacent sentences of the target revision, and sequence labeling to utilize the interdependent revisions. 
Inspired by this work, we 
propose a new approach to extract longer context information. 

Prior studies of revision quality in writing 
have not considered  feedback students receive before revision when defining an annotation scheme \cite{tan2014l,afrin2018improvement}, or have not explored the benefit of using  feedback during 
classification~\cite{afrin2020RER}. 
We leverage both pre-defined AWE feedback messages and free form peer feedback 
in identifying desirable revisions.

Previous studies have explored revision generation for argument writing task~\cite{ito-etal-2019-diamonds} and paraphrase generation tasks ~\cite{mu-lim-2022-revision}. However, state-of-the-art language models are not leveraged for revision classification task. 
The pre-trained Bidirectional Encoder Representations from Transformer (BERT)~\cite{devlin2019BERT} model has shown to be effective in various NLP models including sentence classification and sentence-pair classification. BERT has also produced excellent results in various argument mining tasks~\cite{chakrabarty2019-ampersand, reimers2019sbd, ghosh-etal-2021-laughing}. 
In this work, we leverage the standard pre-trained BERT model (bert-based-uncased) ~\cite{devlin2019BERT} to create the model for our revision classification task.

\section{Data and Resources}
\label{section: data and resources}

Our data consists of three corpora of paired
drafts of argumentative essays, written in response to a prompt and revised in response to feedback. 
A comparison of the data is shown in Table~\ref{table:datasets}. The diversity of the corpora along multiple dimensions helps ensure the utility of our proposed models.

The {\it elementary} school 
students wrote Draft1 about an article on 
a project in Kenya, then received AWE system feedback  focused  on students' use of text evidence and reasoning 
(selected based on 
automatic scoring). An example of the feedback messages is shown in Table~\ref{table:feedback examples}.
All essay pairs were later graded 
on a scale from 0 to 3 to indicate improvement from Draft1 to Draft2 
in line with the feedback  (kappa = 0.77) ~\cite{wang2020eRevise}. 

The {\it high-school} 
students wrote Draft1 in response to a prompt about Dante's Inferno \cite{zhang2015l}, then received peer feedback along 6 rubric dimensions (e.g., evidence, organization, etc.). We only utilize feedback about evidence in this work (shown in Table~\ref{table:feedback examples}), because it is closely related to the revisions we are considering. 
Drafts 1 and 2 of each high-school essay were separately graded 
by expert graders. 
We create an improvement score for each essay pair, calculated  as the difference of the holistic 
score between drafts. 

%
%
%
%


%
%
%

\begin{table*}[t]
\centering
\begin{tabular}{|l|p{1.3cm}|p{2.7cm}|p{2.2cm}|p{5.5cm}|}
\hline
 & Desirable & Undesirable & Desirable & Undesirable \\ 
  & Evidence & Evidence & Reasoning & Reasoning \\ \hline
\multirow{2}{*}{Elementary} 
& \multirow{4}{*}{Relevant} 
& \multirow{4}{2.7cm}{Irrelevant+Repeat +Non-Text-Based + Minimal} 
& LCE + Paraphrase
& Not-LCE + Generic + Commentary + Minimal \\ \cline{1-1}\cline{4-5}

High-school  & & & \multirow{2}{*}{LCE} & \multirow{2}{5cm}{Paraphrase+ Not-LCE+ Generic + Commentary+ Minimal}\\ 
College & & & & \\ \hline
\end{tabular}
\caption{Desirable and Undesirable revision mapping. 
}\label{table: rer mapping to desirable}
\end{table*}

The {\it college} essays were written by 60  students on technology proliferation 
\cite{zhang2017hh}. 
Students received general feedback after Draft1, then revised to create Draft2, then revised again without  
any further textual\footnote{Feedback was given using  AWE interface visualizations.} feedback to create Draft3. 
Drafts 2 and 3 were later graded by experts based on a rubric. 
We create a binary improvement score for each essay pair, calculated as 1 if Draft3 improved compared to Draft2, -1 otherwise.

\begin{table*}[t]
\centering
\begin{tabular}{|l|l|c|c|c|c|c|c|c|}
\hline
&	& &	\multicolumn{3}{c|}{Before Augmentation}	&	\multicolumn{3}{c|}{After Augmentation}		\\	\cline{3-9}
&	& N 	&	Desirable	&	Undesirable	&	Total 	&	Desirable	&	Undesirable	& 	Total\\	\hline
\multirow{2}{*}{Evidence} &Elementary	    & 143		&	239	&	147	&	386 	&	4658	&	2946 &	7604	\\	
&High-school   & 47		&	80	&	30	&	110 	&	1168	&	511	&	1679\\	\hline
\multirow{3}{*}{Reasoning} &Elementary	& 143 	&	186	&	203 &	389		&	3881	&	3844 &	7725	\\	
&High-school	& 47 	&	202	&	185	&	387 	&	2963	&	2817 &	5780	\\	
&College	& 60 	&	114	&	93	&	207 	&	3186	&	2329	&	5515 \\	\hline
\end{tabular}
\caption{Statistics for number of revisions in each corpus. Average number of revisions over 10-fold cross-validation is shown after data augmentation (N = \#Student).}
\label{table: desirable revisions data statistics}
\end{table*}
For all corpora, sentences from the two drafts were aligned manually based on semantic similarity.
Aligned sentences represent one of four operations between drafts -- no change, modification, sentence  deleted from Draft1, sentence added to Draft2.
Each pair of changed aligned sentences was then extracted as a \textit{revision} (rows 3, 5 and 9 in Table \ref{table: example revision}) and annotated for its {\it purpose} (revise reasoning, evidence, and reasoning in rows 3, 5 and 9, respectively). Kappa of the purpose annotation was 0.753~\cite{afrin2020RER}. 
From among the full set of annotations, we only use  evidence and reasoning revisions for the current study because they are the most frequent for elementary and high-school 
data\footnote{1475 revisions were extracted from elementary-school data. Other 700 revisions (claim, word-usage, grammar mistakes, etc.) are not considered due to low frequency. 
1269 revisions were extracted from high-school data. Other 772 revisions are not considered due to low frequency.}. Due to low frequency of evidence revisions, we only use reasoning revisions for college data.

Finally, to  understand how students revise evidence and reasoning, whether their revisions were desirable, and whether desirable revisions relate to  measures of essay improvement, 
we then applied the evidence and reasoning revision categorization scheme developed in ~\cite{afrin2020RER}.
In this scheme, revisions related to evidence are characterized by five codes -- Relevant, Irrelevant, Repeat evidence, Non-text based, and Minimal.
Reasoning revisions are characterized by  six codes -- Linked claim-evidence (LCE), Not LCE, Paraphrase evidence, Generic, Commentary, and Minimal. 
The annotation was done by an expert familiar with the coding scheme (Cohen's kappa in a previous study was 0.833 for evidence and 0.719 for reasoning).








\textbf{Labeling Desirable Revisions.} In this paper, we abstract the evidence and reasoning revision annotations described above into two new categories - {\it desirable} revision and {\it undesirable} revision. The mapping is shown in Table~\ref{table: rer mapping to desirable}. Desirable revisions are those that have hypothesized utility in improving the essay after revision, and are 
encouraged by the writing task. 
Given a different writing task with different feedback messages, different categories may be desirable in improving the essay quality.
For our corpus,  
relevant evidences are desirable because they support a claim in the essay. All the other categories of evidence revisions are combined as undesirable. 
For reasoning revisions, LCE and paraphrase reasoning
are combined as desirable for the elementary-school data\footnote{Paraphrase is encouraged by the writing task.}. On the other hand,  only LCE is a desirable reasoning revision for the high-school and college data. The rest of the reasoning revisions are combined as undesirable. Table~\ref{table: desirable revisions data statistics} shows the number of desirable and undesirable revisions for each corpus~\footnote{See Appendix~\ref{sec: appendix A} for more data distributions.}. We did not combine evidence and reasoning revisions, because 
the schema to label each is different.

\textbf{Extracting Context.}  
We use two methods to extract context of the target revision, {\it simple context} (\textbf{SC})  and {\it longer context} (\textbf{LC}). Following \citet{zhang2016l-context}, we only focus on the sentences before and after the target revision to extract simple context. For example, simple context for the 3rd revision in Table~\ref{table: example revision} consists of sentence 2 and 4 from the revised draft. 
For longer context, we introduce a new method that considers all the sentences that are revised around the target sentence until we find a sentence that is not changed. This makes sure that the context window will have text extracted from both drafts. For example in Table~\ref{table: example revision}, sentence 3 will not have any context from Draft1 using the simple context method.
But with longer context, sentences 1 to 4 from the original draft will be considered as context1 from Draft1; sentences 1 to 4 from revised draft will be considered as context2 from Draft2.
The length of the context will vary depending on the number of revisions within the window. For example, context1 for sentence 3 consists of 2 sentences from Draft1 (1 and 4, 2 was added) while sentence 5 had 3 (4, 6, and 7).


%
%
%
%
\section{Predicting Revision Desirability}
\label{section: predicting revision desirability}
In this section, we describe the models for automatically classifying desirable revisions.
First, we describe a data augmentation process to increase the training data. Then we describe a  model to identify revision desirability, 
and extend it to use context and the feedback information. 
We setup our models to answer the following research questions:

\textbf{RQ1:} Is the context of the revision predictive of revision desirability?

\textbf{RQ2:} Is the feedback received before revising the essay predictive of revision desirability?

\textbf{RQ3:} Do the context and feedback together boost the identification of desirable revision?

\subsection{Data Augmentation}


Our limited amount of revision data is not suitable to experiment with various state-of-the-art machine learning and deep learning models. To generate more training examples, we use a customized version of the synonym replacement (SR) data augmentation strategy -- randomly pick a word from the sentence and replace it with a synonym~\cite{Wei2019EDA:AugTasks}. For each sentence, we replaced one random word with its synonyms but did not consider multiple words at the same time to preserve the hand-annotated revision categories. We ignored stop words, selected words that are more than length of 5 characters, and used maximum 5 synonyms per word to limit the number of data generated. The synonyms are extracted from the Synset from WordNet lexical database from Natural Language Toolkit (NLTK) in Python~\cite{NLTK}, e.g.,  
the word `achieve' in sentence 5 of Table~\ref{table: example revision} can be replaced by `accomplish'. Then the augmented new revision is added as a training instance. 
The last three columns in Table~\ref{table: desirable revisions data statistics} show the average number of revisions after  augmentation.

\subsection{Models}
\begin{figure}[!h]
\centering
\includegraphics[width=0.7\linewidth]{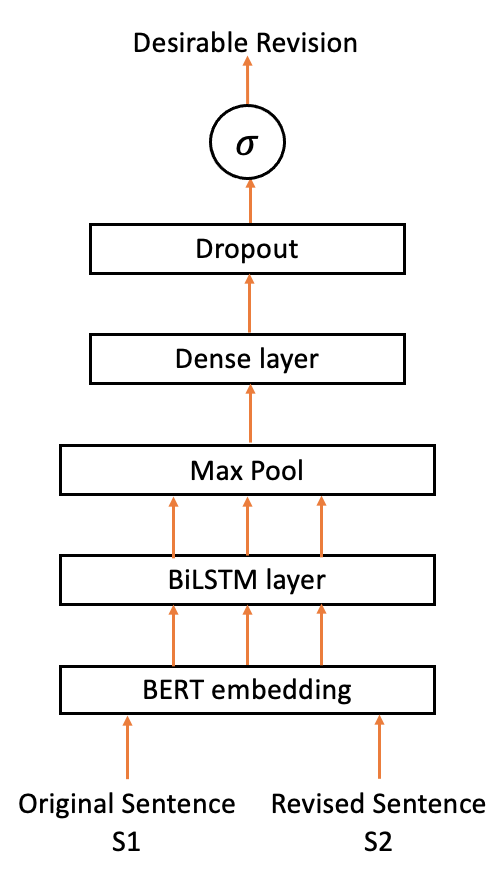}
\caption{Our model M architecture.} 
\label{fig:BERTcls}
\end{figure}

Figure~\ref{fig:BERTcls} shows the neural network model used in this study ({\bf Model M}). 
We used the pre-trained `bert-based-uncased' from Keras Huggingface Library~\cite{devlin2019BERT,wolf-etal-2020-transformers} and encode our revision sentence pair using BERT encoder.
After encoding, we use a BiLSTM layer and a Dense layer to build our neural network  model using the Keras library~\cite{chollet2015keras}.  This architecture  allows easy incorporation of context and feedback as direct inputs, as discussed below.




Bidirectional Long Short Term Memory networks (BiLSTM) has been used in revision classification~\cite{anthonio-roth-2020-learn} in addition to various sentence-pair modeling and sentence classification tasks~\cite{vlad-etal-2019-sentence,he-lin-2016-pairwise} etc. \citet{vlad-etal-2019-sentence} used a BERT-BiLSTM capsule model with additional dense layers with dropout. 
Following these works, we add a BiLSTM layer after extracting the embedding from BERT to process the input sequences.\footnote{We also experimented with simpler neural nets  (e.g., no BiLSTM layer) as our core proposed model, but they did not perform better than 
model {\bf M}.}  
We used a dropout and recurrent dropout rate of 0.1. To down-sample the output representation from the BiLSTM, we take the maximum value over the time dimension using the GlobalMaxPool1D~\cite{chollet2015keras}.

To improve  performance while still keeping the model simple, we add a dense layer after BiLSTM with `relu' as the activation function~\cite{javid2021ARD}. 
In order to make the model robust to overfitting, we add a dropout layer with rate 0.2.
The output is then passed to the final output dense layer with 1 neuron. Since this is a binary classification task, we use `Sigmoid' as the activation function.

\begin{table*}[t]
\centering
\begin{tabular}{|l|c|c|c|c|c|}
\hline	

 		&	\multicolumn{2}{c|}{Elementary} &	\multicolumn{2}{c|}{High-school}	&	College	\\	\hline	

 Model		&	Evidence	&	Reasoning &	 Evidence	&	Reasoning		&	Reasoning	\\	\hline	
	LogR	&	0.469	&	0.537	&	0.470	&	0.495 &	0.462\\
	M	&	0.569	&	0.597	&	0.446	&	0.649   &	0.613 \\
	 +SC	&	0.548	&	0.611	&	0.489	&	\textbf{0.679}  &	0.545\\
	+LC	&	0.574	&	0.627	&	0.474	&	0.665   &	\textbf{0.634}\\
	+F	&	0.570	&	0.639	&	0.452	&	0.652   & -- \\
	+LC\&F &	\textbf{0.587}	&	\textbf{0.649}	&	\textbf{0.521}	&	 0.664  & --\\	\hline


\end{tabular}
\caption{Intrinsic evaluation: average unweighted f1-score over 10-fold cross-validation. 
Best 
are marked bold.}
\label{table: intrinsic}
\end{table*}

We tune the model using Adam optimizer with learning rate \{$1e^{-3}$, $1e^{-4}$, $1e^{-5}$\} and batch size \{16, 32, 64\} using a validation set of 2000 instances extracted from the elementary evidence augmented data. Finally, we select the learning rate at $1e^{-3}$ and batch size 16, and apply the same to all data. 
The hidden layer size is set to 64. There were 434,817 trainable parameters in the model. 

\textbf{Context Model.} 
In this model, in addition to the revision we also provide the context1 from Draft1 and context2 from Draft2 as input to the model to answer \textbf{RQ1}. Since BERT cannot handle more than 512 tokens and our context can be long in some cases, we did not concatenate contexts from two drafts before encoding. First, we encode each context from each draft using the BERT encoder and extract the embedding. Then the context1 and context2 embeddings are concatenated with the revision input in the order of [revision pair, context1, context2]. Then the concatenated embedding is sent to the BiLSTM layer. There is no change in the following layers. When the context is longer than 512 tokens, it is truncated from the end. ~\footnote{No truncation was needed for high-school data. For elementary school, about 9\% and 4\% contexts were deleted for evidence and reasoning, respectively.}

\textbf{Feedback Model.} 
To answer \textbf{RQ2},  we use feedback information to predict revision desirability. We first concatenate all the sentences from the feedback messages. 
Then we encode the whole feedback message using BERT encoder and extract the embedding. The embedding is then concatenated with the input revision from the baseline model in the order of [revision pair, feedback] and sent to the BiLSTM layer.
Feedback messages longer than 512 tokens are truncated from the end.~\footnote{No truncation was needed for elementary data. For high-school, feedback messages were truncated for 55\% of students.}

\textbf{Context \& Feedback Model. }
We also experiment with context and feedback together to answer \textbf{RQ3}. We encode  context and feedback 
as we did in the previous models. The embeddings are then concatenated in the order  [revision pair, context1, context2, feedback] and sent to the BiLSTM layer. 

\textbf{Baseline Model.} We compare our models with a simple model used in prior work that uses logistic regression (LogR)~\cite{afrin2020RER} using GloVe word2vec~\cite{pennington2014glove} features for revision classification. 

\section{Results and Evaluation}

\subsection{Intrinsic Evaluation} 
In our intrinsic evaluation (see Table~\ref{table: intrinsic}), we compare whether context and/or feedback model performance improves over the proposed model {\bf M} in terms of average unweighted 
F1-score~\footnote{See Appendix~\ref{sec: appendix A} for more results.}, 
over 10-folds of cross-validation. Without augmentation, our model does not learn at all from the very small amount of data, hence we only report results using augmented data. Augmentation is done at each fold on the training instances. Test instances are kept original, no augmentation applied. We ran the model 10 epochs for each fold.

First, we compare  model \textbf{M} and its extensions with the LogR baseline. We see that {\bf M} improved over LogR for all cases except high-school evidence classification. Similarly, {\bf M} plus context and/or feedback   improved over LogR in all  cases except with feedback for high-school evidence.

To answer \textbf{RQ1}, we look at the results of the context model and see that 
our proposed longer context representation ({\bf LC}) always improved over {\bf M} (no context), 
which is not true for simple context ({\bf SC}). 
For elementary data, \textbf{LC} performed better than \textbf{SC}, 
while for high-school data, \textbf{SC} performed better than \textbf{LC}. Recall that for high-school data, we did not truncate any context, which means students did not make multiple consecutive revisions frequently. This could explain why \textbf{SC} was better for high-school data. For college data, \textbf{SC} did not improve over \textbf{M}, but \textbf{LC} showed the best performance.

\begin{table*}[!tbh]
\centering

\begin{tabular}{|l|c|c|c|c|c|c|c|c|c|c|}
\hline		
		&	\multicolumn{4}{c|}{Elementary (N=143)}		&	\multicolumn{4}{c|}{High-school (N=47)}		&	\multicolumn{2}{c|}{College (N=60)}\\	\hline
		&	\multicolumn{2}{c|}{Evidence}		&	\multicolumn{2}{c|}{Reasoning} &	\multicolumn{2}{c|}{Evidence}		&	\multicolumn{2}{c|}{Reasoning} &	\multicolumn{2}{c|}{Reasoning}		\\	\hline
		&	D	&	U	&	D	&	U  &	D	&	U	&	D	&	U &	D	&	U	\\	\hline
    Gold	&	0.200*	&	0.039	&	0.450*	&	-0.022	&	0.391*	&	0.040	&	0.351*	&	0.272   &    0.029	    &	-0.131	\\	
    LogR	& 0.112	&	0.182*	&	\textbf{0.231*}	&	0.226*	& 0.229	&	0.240	&	\textbf{0.371*}	&	0.207	&  	0.030	    &	-0.095	\\	
	M	&	0.156	&	0.106	&	\textbf{0.339*}	&	0.114	&	\textbf{0.321*}	&	0.156	&	0.249	&	0.396*	&	0.039       &	-0.181 \\
	+SC	&	0.137	&	0.137	&	\textbf{0.321*}	&	0.093	&	\textbf{0.350*}	&	0.025	&	\textbf{0.335*}	&	0.307*	&	-0.016	&	-0.123	\\	
	+LC	&	0.152	&	0.084	&	\textbf{0.422*}	&	-0.039	& \textbf{0.366*}	&	-0.030	&	\textbf{0.407*}	&	0.257	&	0.083	&	-0.246	\\	
	+F	&	0.125	&	0.162	&	\textbf{0.360*}	&	0.080	& \textbf{0.323*}	&	0.090	&	\textbf{0.327*}	&	0.322*  &   -- & --	\\	
	+LC\&F	&	0.139	&	0.117	&	\textbf{0.381*}	&	0.041	& \textbf{0.354*}	&	-0.064	&	\textbf{0.406*}	&	0.239	&   --  &   --   \\	\hline

	
\end{tabular}


	
\caption{Extrinsic evaluation: significant correlations using predicted desirability that are consistent with using gold labels are marked bold (*  p$<.05$, N = \#Students, D: Desirable, U: Undesirable).}
\label{table: extrinsic evaluation}
\end{table*}
To answer \textbf{RQ2}, 
the results of the feedback model (\textbf{F}) in Table~\ref{table: intrinsic} 
show that while  \textbf{F}  did improve over \textbf{M} for each task,  in most cases the increase is  low.  
Desirable reasoning classification for elementary-school data had the most benefit using the feedback. 
This could be because every elementary-school student was specifically asked to provide more details or explain their evidence. 
For high-school data, although \textbf{F}  improved over \textbf{M}, it did not improve over LogR  for evidence.~\footnote{No feedback available for college data Draft2 and Draft3.}

To answer \textbf{RQ3}, we only consider longer context and feedback messages (\textbf{LC\&F}). 
As shown in Table~\ref{table: intrinsic}, the \textbf{LC\&F} model always improved model \textbf{M}'s performance and has the best performance except high-school reasoning revision. This indicates that feedback messages were most helpful when combined with the context,  especially for elementary-school reasoning revisions where the performance increased more than 0.05 points. This could be  because students did not receive feedback  at sentence-level; instead, the feedback is usually about specific areas of the essay or about the argumentative structure of the essay. Hence, when combined with the context, it helps the model to capture a better picture. 

\subsection{Extrinsic Evaluation}

To confirm that revision desirability is indeed related to the essay improvement scores described in Section~\ref{section: data and resources}, we calculated the Pearson correlation between the frequency of desirable and undesirable revisions  (gold annotations) to improvement score.
For extrinsic evaluation, 
we then replicate the correlation calculation for the predicted labels to see if the frequency of predicted desirable revisions are still correlated to the essay improvement. Table~\ref{table: extrinsic evaluation} shows the gold and predicted correlations.

Model \textbf{M} showed to be consistent with Gold annotations for elementary reasoning and high-school evidence prediction. \textbf{M} also showed higher correlation than LogR when it is consistent with Gold.

\begin{table*}[!htb]
\centering
\begin{tabular}{|p{3cm}|p{5.4cm}|c|c|c|c|c|c|}
\hline	
Original Draft & Revised Draft	 &	Gold	&	M	&	+SC & +LC	&	+F	&	+LC\&F	\\	\hline \hline
&They can do that by assuring that the people of Sauri, Kenya have food, water, liter, and a place to stay.	 &	D R	&	U	& U &	D	&	U	&	D	\\	\hline

We think \$5 dollars isn't that much money but they live in poverty. & We think \$5 dollars isn't that much money but they live in situations where \$5 is a weeks worth of money.	 &	D E	&	D	& U &	U	&	D	&	U	\\	\hline

& They had water, food, electricity, supplies, medicine, and simple things.	 &	U E	&	D	& U &	U	&	D	&	U	\\

\hline

\end{tabular}
\caption{Revision examples with gold and predicted labels. D: Desirable, U: Undesirable, E: Evidence, R: Reasoning}
\label{table: predicted examples}
\end{table*}

%
%
%
Overall, the number of desirable revisions predicted by \textbf{LC} showed the highest R values. 
While we do not expect the models to have higher correlations than the gold annotations, \textbf{LC} did in one case (desirable reasoning prediction for high-school data). 
Gold annotations did not show significant negative correlations to undesirable revisions. This is because the scoring rubrics typically did not penalize for revisions that did not improve the essay, as long as revising didn't make the essay worse. \textbf{LC} also did not show any significant correlation to undesirable revisions. Unexpectedly, \textbf{SC} did in one case (undesirable reasoning for high-school). 


Model \textbf{F} similarly yielded significant positive correlation with desirable revisions and had higher correlations than model \textbf{M}. 
In most cases Model \textbf{F} is consistent with Gold annotations, except for undesirable reasoning revisions for high-school data.

Model \textbf{LC\&F} also showed higher significant correlation for the predicted labels compared to Model \textbf{M}. However, unlike the intrinsic evaluation it does not show us the best performance. 

Unfortunately, we did not see any significant correlation for the college data. But in most cases, desirable revisions showed positive sign, while undesirable revisions showed negative sign.



\section{Qualitative Analysis}
\label{section: discussion}

In order to better understand the model predictions, in Table~\ref{table: predicted examples} we compare gold and predicted labels for a few example revisions.
The first example (taken from  Table~\ref{table: example revision}) is predicted as desirable whenever longer context information was available. Otherwise, it is wrongly predicted as undesirable. Looking at this revision (sentence 3) and its context from Table~\ref{table: example revision}, we can see that sentence 3 mentions about the `people',  
`food, water, liter, and a place to stay'. The context 
mention `people', `health care' and `quality of life'. We think those phrases helped the context model to identify this example as desirable. However, although feedback messages asked to `explain the evidence', the feedback model was not successful in identifying this as desirable.

The second example is a desirable evidence predicted as undesirable by context and desirable by the feedback model. The AWE feedback asked the student to use more evidence and add details. 
We think the feedback model tied the extra information in the modified sentence to what was asked for. 


The last example is an undesirable evidence predicted correctly only by the models using context information. 
Although the example text resembles a desirable evidence, it is actually undesirable because it was repeated. Obviously, the model needed context to identify that it is a repeated evidence. 
%
%
%
%







%
%
%
%
\section{Conclusion}
In this study, we presented new models for the automatic identification of desirable revisions in three corpora of argumentative writing varying in writer's level of expertise, source of feedback, and grading rubrics. We presented a new method of extracting context from essay revisions. Using intrinsic and extrinsic evaluation we showed that models using the context information performed best in identifying desirable revisions. We also studied the use of feedback messages
received by  students to predict desirable revisions. To the best of our knowledge this is the first model to use feedback information to analyze student revision. Our experiments showed that feedback information also helped improve  classifier performance, particularly when used with context. We have released the college data annotated with revision desirability. It can be downloaded from this link: \url{https://petal-cs-pitt.github.io/data.html}. The code is also available from here: \url{https://github.com/tazin-afrin/desirable-revision-classification}

\section*{Acknowledgements}
The research reported here was supported, in whole
or in part, by the National Science Foundation (NSF) grant 1735752 and 2202347 to the University of Pittsburgh. We would like to thank the anonymous reviewers for taking the time to review our paper and provide us with detailed feedback. We would also like to thank the members of the PETAL lab for their valuable feedback. The opinions expressed are those of the authors and do not represent the views of the Institute.

\section*{Discussion of Limitations}

Our use of both context and feedback could be enhanced in future work. First, we sometimes needed to truncate context or feedback from the end, which may remove useful information. In the future, we plan to use other transformer architectures capable of handling longer sequences (e.g., Longformer~\cite{Beltagy2020Longformer}). 
Second, while our proposed method of extracting longer context enables the use of variable length context windows, 
our method does not guarantee that the context will include the major claim. Since evidence and reasoning 
are most effective when used to support a  claim, their revision desirability  might depend on the essay's claim.
Third, since the feedback received by students 
was largely framed at the essay-level, we did not attempt to connect the messages with specific sentence revisions.  Such modeling could potentially improve feedback performance.

Additional limitations include that our classifier input was based on perfect alignment of the sentences in the essay drafts and used gold evidence and reasoning revision purpose labels. An end-to-end system would have lower performance due to errors propagated from alignment and purpose classification. 
Our data is also limited 
in that essays are all of an argumentative writing style and annotated for  only  two types of content revisions. Also, the corpus is small.  Although, we used simple augmentation to generate enough data to experiment with complex learning models, in the future we plan to explore other options for data augmentation. We also would like to use similar argumentative essays to fine-tune the BERT architecture.  

\section*{Ethical Considerations}

All corpora were collected under protocols approved by an institutional review board, including that the data is not publicly available, except the college data. While the breach of private student information from the elementary and high school data will thus not pose any ethical concern, other researchers can not replicate our results for those data. However, since the college data with its purpose annotations was already made available by the original researchers, our new desirability annotations can be released upon acceptance of this study. The claims of the paper match the experimental results and the results can be hypothesized to generalize. 
In the future, the proposed models may be incorporated into AWE systems for student writers. While  identifying and providing feedback on revision desirability will be helpful to students in improving their writing, there is the risk that the system might sometimes provide poor advice based on incorrect model classifications. 
Since the dataset is still fairly small after data augmentation, it is possible that the model may learn biased representation of the revisions (e.g., always predict longer revisions with more information as desirable).

\bibliography{anthology}

\begin{thebibliography}{37}
\expandafter\ifx\csname natexlab\endcsname\relax\def\natexlab#1{#1}\fi

\bibitem[{Afrin and Litman(2018)}]{afrin2018improvement}
Tazin Afrin and Diane Litman. 2018.
\newblock Annotation and classification of sentence-level revision improvement.
\newblock In \emph{Proceedings of the Thirteenth Workshop on Innovative Use of
  NLP for Building Educational Applications}, pages 240--246, New Orleans,
  Louisiana.

\bibitem[{Afrin and Litman(2019)}]{afrin2019EditorRole}
Tazin Afrin and Diane~J. Litman. 2019.
\newblock Identifying editor roles in argumentative writing from student
  revision histories.
\newblock In \emph{Artificial Intelligence in Education - 20th International
  Conference, {AIED} 2019, Chicago, IL, USA, June 25-29, 2019, Proceedings,
  Part {II}}, volume 11626 of \emph{Lecture Notes in Computer Science}, pages
  9--13. Springer.

\bibitem[{Afrin et~al.(2020)Afrin, Wang, Litman, Matsumura, and
  Correnti}]{afrin2020RER}
Tazin Afrin, Elaine~Lin Wang, Diane Litman, Lindsay~Clare Matsumura, and
  Richard Correnti. 2020.
\newblock Annotation and classification of evidence and reasoning revisions in
  argumentative writing.
\newblock In \emph{Proceedings of the Fifteenth Workshop on Innovative Use of
  NLP for Building Educational Applications}, Seattle, Washington, USA
  (Remote).

\bibitem[{Anthonio and Roth(2020)}]{anthonio-roth-2020-learn}
Talita Anthonio and Michael Roth. 2020.
\newblock \href {https://doi.org/10.18653/v1/2020.coling-main.117} {What can we
  learn from noun substitutions in revision histories?}
\newblock In \emph{Proceedings of the 28th International Conference on
  Computational Linguistics}, pages 1359--1370, Barcelona, Spain (Online).
  International Committee on Computational Linguistics.

\bibitem[{Beigman~Klebanov and Madnani(2020)}]{beigman2020AWE}
Beata Beigman~Klebanov and Nitin Madnani. 2020.
\newblock \href {https://doi.org/10.18653/v1/2020.acl-main.697} {Automated
  evaluation of writing {--} 50 years and counting}.
\newblock In \emph{Proceedings of the 58th Annual Meeting of the Association
  for Computational Linguistics}, pages 7796--7810, Online. Association for
  Computational Linguistics.

\bibitem[{Beltagy et~al.(2020)Beltagy, Peters, and
  Cohan}]{Beltagy2020Longformer}
Iz~Beltagy, Matthew~E. Peters, and Arman Cohan. 2020.
\newblock Longformer: The long-document transformer.
\newblock \emph{arXiv:2004.05150}.

\bibitem[{Bird et~al.(2009)Bird, Klein, and Loper}]{NLTK}
Steven Bird, Ewan Klein, and Edward Loper. 2009.
\newblock \emph{Natural language processing with Python: analyzing text with
  the natural language toolkit}.
\newblock " O`Reilly Media, Inc.".

\bibitem[{Bronner and Monz(2012)}]{bronner2012m}
Amit Bronner and Christof Monz. 2012.
\newblock \href {http://dl.acm.org/citation.cfm?id=2380816.2380860} {User edits
  classification using document revision histories}.
\newblock In \emph{Proceedings of the 13th Conference of the European Chapter
  of the Association for Computational Linguistics}, EACL '12, pages 356--366,
  Avignon, France. Association for Computational Linguistics.

\bibitem[{Chakrabarty et~al.(2019)Chakrabarty, Hidey, Muresan, McKeown, and
  Hwang}]{chakrabarty2019-ampersand}
Tuhin Chakrabarty, Christopher Hidey, Smaranda Muresan, Kathy McKeown, and
  Alyssa Hwang. 2019.
\newblock {AMPERSAND}: Argument mining for {PERS}u{A}sive o{N}line discussions.
\newblock In \emph{Proceedings of the 2019 Conference on Empirical Methods in
  Natural Language Processing and the 9th International Joint Conference on
  Natural Language Processing (EMNLP-IJCNLP)}, pages 2933--2943, Hong Kong,
  China. Association for Computational Linguistics.

\bibitem[{Chollet et~al.(2015)}]{chollet2015keras}
Francois Chollet et~al. 2015.
\newblock Keras, https://github.com/fchollet/keras, [online; accessed
  10-10-2021].

\bibitem[{Daxenberger and Gurevych(2012)}]{daxenberger2012g}
Johannes Daxenberger and Iryna Gurevych. 2012.
\newblock A corpus-based study of edit categories in featured and non-featured
  wikipedia articles.
\newblock In \emph{Proceedings of the 24th International Conference on
  Computational Linguistics}, COLING '12, pages 711--726, Mumbai, India.

\bibitem[{Daxenberger and Gurevych(2013)}]{daxenbergerG13g}
Johannes Daxenberger and Iryna Gurevych. 2013.
\newblock Automatically classifying edit categories in wikipedia revisions.
\newblock In \emph{Proceedings of the 2013 Conference on Empirical Methods in
  Natural Language Processing}, EMNLP '13, pages 578--589, Seattle, Washington,
  USA. Association for Computational Linguistics.

\bibitem[{Debnath and Roth(2021)}]{debnath-roth-2021-computational}
Alok Debnath and Michael Roth. 2021.
\newblock \href {https://doi.org/10.18653/v1/2021.eacl-srw.5} {A computational
  analysis of vagueness in revisions of instructional texts}.
\newblock In \emph{Proceedings of the 16th Conference of the European Chapter
  of the Association for Computational Linguistics: Student Research Workshop},
  pages 30--35, Online. Association for Computational Linguistics.

\bibitem[{Devlin et~al.(2019)Devlin, Chang, Lee, and
  Toutanova}]{devlin2019BERT}
Jacob Devlin, Ming-Wei Chang, Kenton Lee, and Kristina Toutanova. 2019.
\newblock {BERT}: Pre-training of deep bidirectional transformers for language
  understanding.
\newblock In \emph{Proceedings of the 2019 Conference of the North {A}merican
  Chapter of the Association for Computational Linguistics: Human Language
  Technologies, Volume 1 (Long and Short Papers)}, pages 4171--4186,
  Minneapolis, Minnesota. Association for Computational Linguistics.

\bibitem[{Du et~al.(2022)Du, Raheja, Kumar, Kim, Lopez, and
  Kang}]{du-etal-2022-understanding-iterative}
Wanyu Du, Vipul Raheja, Dhruv Kumar, Zae~Myung Kim, Melissa Lopez, and Dongyeop
  Kang. 2022.
\newblock \href {https://aclanthology.org/2022.acl-long.250} {Understanding
  iterative revision from human-written text}.
\newblock In \emph{Proceedings of the 60th Annual Meeting of the Association
  for Computational Linguistics (Volume 1: Long Papers)}, pages 3573--3590,
  Dublin, Ireland. Association for Computational Linguistics.

\bibitem[{Faigley and Witte(1981)}]{faigley1981w}
Lester Faigley and Stephen Witte. 1981.
\newblock Analyzing revision.
\newblock \emph{College Composition and Communication}, 32(4):400--414.

\bibitem[{Ghosh et~al.(2021)Ghosh, Shrivastava, and
  Muresan}]{ghosh-etal-2021-laughing}
Debanjan Ghosh, Ritvik Shrivastava, and Smaranda Muresan. 2021.
\newblock \href {https://aclanthology.org/2021.eacl-main.171} {{``}laughing at
  you or with you{''}: The role of sarcasm in shaping the disagreement space}.
\newblock In \emph{Proceedings of the 16th Conference of the European Chapter
  of the Association for Computational Linguistics: Main Volume}, pages
  1998--2010, Online. Association for Computational Linguistics.

\bibitem[{He and Lin(2016)}]{he-lin-2016-pairwise}
Hua He and Jimmy Lin. 2016.
\newblock \href {https://doi.org/10.18653/v1/N16-1108} {Pairwise word
  interaction modeling with deep neural networks for semantic similarity
  measurement}.
\newblock In \emph{Proceedings of the 2016 Conference of the North {A}merican
  Chapter of the Association for Computational Linguistics: Human Language
  Technologies}, pages 937--948, San Diego, California. Association for
  Computational Linguistics.

\bibitem[{Ito et~al.(2019)Ito, Kuribayashi, Kobayashi, Brassard, Hagiwara,
  Suzuki, and Inui}]{ito-etal-2019-diamonds}
Takumi Ito, Tatsuki Kuribayashi, Hayato Kobayashi, Ana Brassard, Masato
  Hagiwara, Jun Suzuki, and Kentaro Inui. 2019.
\newblock \href {https://aclanthology.org/W19-8606} {Diamonds in the rough:
  Generating fluent sentences from early-stage drafts for academic writing
  assistance}.
\newblock In \emph{Proceedings of the 12th International Conference on Natural
  Language Generation}, pages 40--53, Tokyo, Japan. Association for
  Computational Linguistics.

\bibitem[{Javid et~al.(2021)Javid, Das, Skoglund, and
  Chatterjee}]{javid2021ARD}
Alireza~M. Javid, Sandipan Das, Mikael Skoglund, and Saikat Chatterjee. 2021.
\newblock A relu dense layer to improve the performance of neural networks.
\newblock \emph{ICASSP 2021 - 2021 IEEE International Conference on Acoustics,
  Speech and Signal Processing (ICASSP)}, pages 2810--2814.

\bibitem[{Kashefi et~al.(2022)Kashefi, Afrin, Dale, Olshefski, Godley, Litman,
  and Hwa}]{kashefi2022}
Omid Kashefi, Tazin Afrin, Meghan Dale, Christopher Olshefski, Amanda Godley,
  Diane Litman, and Rebecca Hwa. 2022.
\newblock \href {https://doi.org/10.1007/s10579-021-09567-z} {Argrewrite v.2:
  an annotated argumentative revisions corpus}.
\newblock \emph{Language Resources and Evaluation}, pages 1574--0218.

\bibitem[{Mu and Lim(2022)}]{mu-lim-2022-revision}
Wenchuan Mu and Kwan~Hui Lim. 2022.
\newblock \href {https://aclanthology.org/2022.tsar-1.6} {Revision for
  concision: A constrained paraphrase generation task}.
\newblock In \emph{Proceedings of the Workshop on Text Simplification,
  Accessibility, and Readability (TSAR-2022)}, pages 57--76, Abu Dhabi, United
  Arab Emirates (Virtual). Association for Computational Linguistics.

\bibitem[{Pennington et~al.(2014)Pennington, Socher, and
  Manning}]{pennington2014glove}
Jeffrey Pennington, Richard Socher, and Christopher~D. Manning. 2014.
\newblock Glove: Global vectors for word representation.
\newblock In \emph{Empirical Methods in Natural Language Processing (EMNLP)},
  pages 1532--1543.

\bibitem[{Reimers et~al.(2019)Reimers, Schiller, Beck, Daxenberger, Stab, and
  Gurevych}]{reimers2019sbd}
Nils Reimers, Benjamin Schiller, Tilman Beck, Johannes Daxenberger, Christian
  Stab, and Iryna Gurevych. 2019.
\newblock Classification and clustering of arguments with contextualized word
  embeddings.
\newblock In \emph{Proceedings of the 57th Annual Meeting of the Association
  for Computational Linguistics}, pages 567--578, Florence, Italy. Association
  for Computational Linguistics.

\bibitem[{Shibani et~al.(2018)Shibani, Knight, and
  Buckingham~Shum}]{shibani2018kb}
Antonette Shibani, Simon Knight, and Simon Buckingham~Shum. 2018.
\newblock Understanding revisions in student writing through revision graphs.
\newblock In \emph{International Conference on Artificial Intelligence in
  Education}, pages 332--336, Cham. Springer International Publishing.

\bibitem[{Skitalinskaya et~al.(2021)Skitalinskaya, Klaff, and
  Wachsmuth}]{skitalinskaya-etal-2021-learning}
Gabriella Skitalinskaya, Jonas Klaff, and Henning Wachsmuth. 2021.
\newblock \href {https://aclanthology.org/2021.eacl-main.147} {Learning from
  revisions: Quality assessment of claims in argumentation at scale}.
\newblock In \emph{Proceedings of the 16th Conference of the European Chapter
  of the Association for Computational Linguistics: Main Volume}, pages
  1718--1729, Online. Association for Computational Linguistics.

\bibitem[{Tan and Lee(2014)}]{tan2014l}
Chenhao Tan and Lillian Lee. 2014.
\newblock A corpus of sentence-level revisions in academic writing: {A} step
  towards understanding statement strength in communication.
\newblock In \emph{Proceedings of the 52nd Annual Meeting of the Association
  for Computational Linguistics}, volume 2: Short Papers, pages 403--408,
  Baltimore, MD, USA.

\bibitem[{Vlad et~al.(2019)Vlad, Tanase, Onose, and
  Cercel}]{vlad-etal-2019-sentence}
George-Alexandru Vlad, Mircea-Adrian Tanase, Cristian Onose, and
  Dumitru-Clementin Cercel. 2019.
\newblock \href {https://doi.org/10.18653/v1/D19-5022} {Sentence-level
  propaganda detection in news articles with transfer learning and
  {BERT}-{B}i{LSTM}-capsule model}.
\newblock In \emph{Proceedings of the Second Workshop on Natural Language
  Processing for Internet Freedom: Censorship, Disinformation, and Propaganda},
  pages 148--154, Hong Kong, China. Association for Computational Linguistics.

\bibitem[{Wang et~al.(2020)Wang, Matsumura, Correnti, Litman, Zhang, Howe,
  Magooda, and Quintana}]{wang2020eRevise}
Elaine~Lin Wang, Lindsay~Clare Matsumura, Richard Correnti, Diane Litman,
  Haoran Zhang, Emily Howe, Ahmed Magooda, and Rafael Quintana. 2020.
\newblock erevis(ing): Students’ revision of text evidence use in an
  automated writing evaluation system.
\newblock \emph{Assessing Writing}, 44:100449.

\bibitem[{Wei and Zou(2019)}]{Wei2019EDA:AugTasks}
Jason Wei and Kai Zou. 2019.
\newblock {EDA: Easy data augmentation techniques for boosting performance on
  text classification tasks}.
\newblock In \emph{EMNLP}, pages 6382--6388.

\bibitem[{Wolf et~al.(2020)Wolf, Debut, Sanh, Chaumond, Delangue, Moi, Cistac,
  Rault, Louf, Funtowicz, Davison, Shleifer, von Platen, Ma, Jernite, Plu, Xu,
  Scao, Gugger, Drame, Lhoest, and Rush}]{wolf-etal-2020-transformers}
Thomas Wolf, Lysandre Debut, Victor Sanh, Julien Chaumond, Clement Delangue,
  Anthony Moi, Pierric Cistac, Tim Rault, Rémi Louf, Morgan Funtowicz, Joe
  Davison, Sam Shleifer, Patrick von Platen, Clara Ma, Yacine Jernite, Julien
  Plu, Canwen Xu, Teven~Le Scao, Sylvain Gugger, Mariama Drame, Quentin Lhoest,
  and Alexander~M. Rush. 2020.
\newblock \href {https://www.aclweb.org/anthology/2020.emnlp-demos.6}
  {Transformers: State-of-the-art natural language processing}.
\newblock In \emph{Proceedings of the 2020 Conference on Empirical Methods in
  Natural Language Processing: System Demonstrations}, pages 38--45, Online.
  Association for Computational Linguistics.

\bibitem[{Writing~Mentor(2016)}]{ets-writing-mentor}
The Writing~Mentor. 2016.
\newblock {ETS} writing mentor, https://mentormywriting.org/, [online; accessed
  02-06-2019].

\bibitem[{Yang et~al.(2017)Yang, Halfaker, Kraut, and Hovy}]{yang2017hkh}
Diyi Yang, Aaron Halfaker, Robert~E. Kraut, and Eduard~H. Hovy. 2017.
\newblock Identifying semantic edit intentions from revisions in wikipedia.
\newblock In \emph{Proceedings of the 2017 Conference on Empirical Methods in
  Natural Language Processing}, EMNLP'17, pages 9--11, Copenhagen, Denmark.
  Association for Computational Linguistics.

\bibitem[{Zhang et~al.(2017)Zhang, Hashemi, Hwa, and Litman}]{zhang2017hh}
Fan Zhang, Homa Hashemi, Rebecca Hwa, and Diane Litman. 2017.
\newblock A corpus of annotated revisions for studying argumentative writing.
\newblock In \emph{Proceedings of the 55th Annual Meeting of the Association
  for Computational Linguistics (Volume 1: Long Papers)}, pages 1568--1578.
  Association for Computational Linguistics.

\bibitem[{Zhang et~al.(2016)Zhang, Hwa, Litman, and B.~Hashemi}]{zhang2016hl}
Fan Zhang, Rebecca Hwa, Diane Litman, and Homa B.~Hashemi. 2016.
\newblock Argrewrite: A web-based revision assistant for argumentative
  writings.
\newblock In \emph{Proceedings of the 2016 Conference of the North American
  Chapter of the Association for Computational Linguistics: Demonstrations},
  pages 37--41, San Diego, California. Association for Computational
  Linguistics.

\bibitem[{Zhang and Litman(2015)}]{zhang2015l}
Fan Zhang and Diane Litman. 2015.
\newblock Annotation and classification of argumentative writing revisions.
\newblock In \emph{Proceedings of the 10th Workshop on Innovative Use of NLP
  for Building Educational Applications}, pages 133--143, Denver, Colorado.
  Association for Computational Linguistics.

\bibitem[{Zhang and Litman(2016)}]{zhang2016l-context}
Fan Zhang and Diane Litman. 2016.
\newblock Using context to predict the purpose of argumentative writing
  revisions.
\newblock In \emph{Proceedings of the 2016 Conference of the North {A}merican
  Chapter of the Association for Computational Linguistics: Human Language
  Technologies}, pages 1424--1430, San Diego, California. Association for
  Computational Linguistics.

\end{thebibliography}
\bibliographystyle{acl_natbib}

\clearpage

\appendix

\section{Appendix A: Additional Results}
\label{sec: appendix A}

\begin{table}[!htb]
\centering
\begin{tabular}{|l|l|rrrr|}
\hline										
Data &  Revision	&	Add	&	Delete	&	Modify	&	Total	\\ \hline	\hline
\multirow{6}{*}{Elementary (N=143)} & Total Evidence	&	265	&	63	&	58	&	386	\\ 	
& Desirable Evidence		&	159	&	50	&	30	&	239	\\ 
& Undesirable	Evidence &   106	&   13	&   28	&   147	\\ \cline{2-6}
& Total Reasoning	&	270	&	59	&	60	&	389	\\ 
& Desirable Reasoning	    &   140	&   28	&   18	&   186	\\ 
& Undesirable	Reasoning    &   130	&   31	&   42	&   203	\\ \hline \hline

\multirow{6}{*}{High-school (N=47)} & Total Evidence	&	93	&	10	&	7	&	110	\\
& Desirable Evidence	&	73	&	7	&	0	&	80	\\
& Undesirable Evidence	&	20	&	3	&	7	&	30	\\ \cline{2-6}
& Total Reasoning	&	324	&	40	&	23	&	387	\\
& Desirable Reasoning	&	184	&	13	&	5	&	202	\\
& Undesirable Reasoning	&	140	&	27	&	18	&	185	\\ \hline \hline

\multirow{6}{*}{College (N=60)}	& Total Evidence	&	25	&	1	&	0	&	26	\\
& Desirable Evidence	&	23	&	1	&	0	&	24	\\
& Undesirable Evidence	&	2	&	0	&	0	&	2	\\ \cline{2-6}
& Total Reasoning	&	191	&	13	&	3	&	207	\\
& Desirable Reasoning	&	104	&	7	&	3	&	114	\\
& Undesirable Reasoning	&	87	&	6	&	0	&	93	\\ \hline
\end{tabular}
\caption{Detailed data distribution.}
\label{table: apdx RER code distribution}
\end{table}


\begin{table}[!h]
\centering
\begin{tabular}{|l|l|l|l|l|l|l|l|}
\hline			
	&		&	\multicolumn{3}{c}{Evidence} &	\multicolumn{3}{|c|}{Reasoning}	\\	\hline	
	&		&	Precision	&	Recall	&	F1-score	&	Precision	&	Recall	&	F1-score	\\	\hline	\hline
Elementary	&	LogR	&	0.510	&	0.519	&	0.469	&	0.572	&	0.573	&	0.537	\\		
	&	M	&	0.587	&	0.587	&	0.569	&	0.613	&	0.609	&	0.597	\\
	&	 +SC	&	0.587	&	0.575	&	0.548	&	0.624	&	0.626	&	0.611	\\		
	&	+LC	&	\textbf{0.640}	&	0.594	&	0.574	&	0.644	&	0.638	&	0.627	\\
	&	+F	&	0.592	&	0.595	&	0.570	&	0.675	&	0.658	&	0.639	\\
	&	+LC\&F	&	0.636	&	\textbf{0.605}	&	\textbf{0.587}	&	\textbf{0.681}   &	\textbf{0.664}   &	\textbf{0.649}	\\	\hline	\hline

High-school	&	LogR	&	0.493	&	0.535	&	0.470	&	0.600	&	0.555	&	0.495	\\		
	&	M	&	0.434	&	0.476	&	0.446	&	0.668	&	0.662	&	0.649	\\
	&	 +SC	&	0.489	&	0.535	&	0.489	&	\textbf{0.701}	&	\textbf{0.690}	&	\textbf{0.679}	\\		
	&	+LC	&	0.480	&	0.502	&	0.474	&	0.681	&	0.673	&	0.665	\\
	&	+F	&	0.469	&	0.480	&	0.452	&	0.668	&	0.663	&	0.652	\\
	&	+LC\&F	&	\textbf{0.554}	&	\textbf{0.549}	&	\textbf{0.521}	&	0.683	&   0.679	&   0.664	\\	\hline	\hline

College	&	LogR	&		&		&		&	0.507*	&	0.514	&	0.462*	\\		
	&	M	&		&		&		&	0.667	&	0.653	&	0.613	\\
	&	 +SC	&		&		&		&	0.593	&	0.593	&	0.545	\\		
	&	+LC	&		&		&		&	\textbf{0.703}	&	\textbf{0.670}	&	\textbf{0.634}	\\	\hline

\end{tabular}
\caption{10-fold cross-validation result for classifying desirable evidence and reasoning, more metrics.}
\label{table: apdx BERT-basic}
\end{table}

\end{document}